\documentclass[10pt,twocolumn,letterpaper]{article}

\usepackage{cvpr}
\usepackage{times}
\usepackage{epsfig}
\usepackage{graphicx}
\usepackage{amsmath}
\usepackage{amssymb}
\usepackage[linesnumbered,ruled,vlined,lined,boxed,commentsnumbered]{algorithm2e}
\usepackage{subfigure}

\usepackage[pagebackref=true,breaklinks=true,letterpaper=true,colorlinks,bookmarks=false]{hyperref}

\usepackage[usenames, dvipsnames]{color}

\cvprfinalcopy 

\def\cvprPaperID{1950} 

\ifcvprfinal\fi
\begin{document}

\title{Fine-grained Categorization and Dataset Bootstrapping using Deep Metric Learning with Humans in the Loop}


\author{Yin Cui$^{1,2}$~~~~~~~~~~~Feng Zhou$^3$~~~~~~~~~~~Yuanqing Lin$^3$~~~~~~~~~~~Serge Belongie$^{1,2}$\\
\small\\
{ $^1$Department of Computer Science, Cornell University~~~~$^2$Cornell Tech~~~~$^3$NEC Labs America} \\
{\tt\small $^{1,2}$\{ycui, sjb\}@cs.cornell.edu~~~~$^3$\{feng, ylin\}@nec-labs.com}
}

\maketitle

\begin{abstract}
Existing fine-grained visual categorization methods often suffer from three challenges: lack of training data, large number of fine-grained categories, and high intra-class vs.\ low inter-class variance.
In this work we propose a generic iterative framework for fine-grained categorization and dataset bootstrapping that handles these three challenges.
Using deep metric learning with humans in the loop, we learn a low dimensional feature embedding with anchor points on manifolds for each category.
These anchor points capture intra-class variances and remain discriminative between classes.
In each round, images with high confidence scores from our model are sent to humans for labeling.
By comparing with exemplar images, labelers mark each candidate image as either a ``true positive" or a ``false positive."
True positives are added into our current dataset and false positives are regarded as ``hard negatives" for our metric learning model.
Then the model is re-trained with an expanded dataset and hard negatives for the next round.
To demonstrate the effectiveness of the proposed framework, we bootstrap a fine-grained flower dataset with 620 categories from Instagram images.
The proposed deep metric learning scheme is evaluated on both our dataset and the CUB-200-2001 Birds dataset.
Experimental evaluations show significant performance gain using dataset bootstrapping and demonstrate state-of-the-art results achieved by the proposed deep metric learning methods.
\end{abstract}





\section{Introduction}
Fine-grained visual categorization (FGVC) has received increased interest from the computer vision community in recent years.
By definition, FGVC, as a sub-field of object recognition, aims to distinguish subordinate categories within an entry-level category.
For example, in fine-grained flower categorization \cite{flower_ox_1, flower_102, flower_nec}, we want to identify the species of a flower in an image, such as ``nelumbo nucifera (lotus flower)," ``tulip" or ``cherry blossom."
Other examples include classifying different types of plants \cite{leafsnap}, birds \cite{branson2010visual, pose_normlized_net}, dogs \cite{stanford_dog}, insects \cite{larios2010haar}, galaxies \cite{dieleman2015rotation, cui2014spatial}; recognizing brand, model and year of cars \cite{stanford_car, xiang2015car, cuhk_car}; and face identification \cite{deepface, facenet}.

\begin{figure}[t]
\begin{center}
\includegraphics[width=1\linewidth]{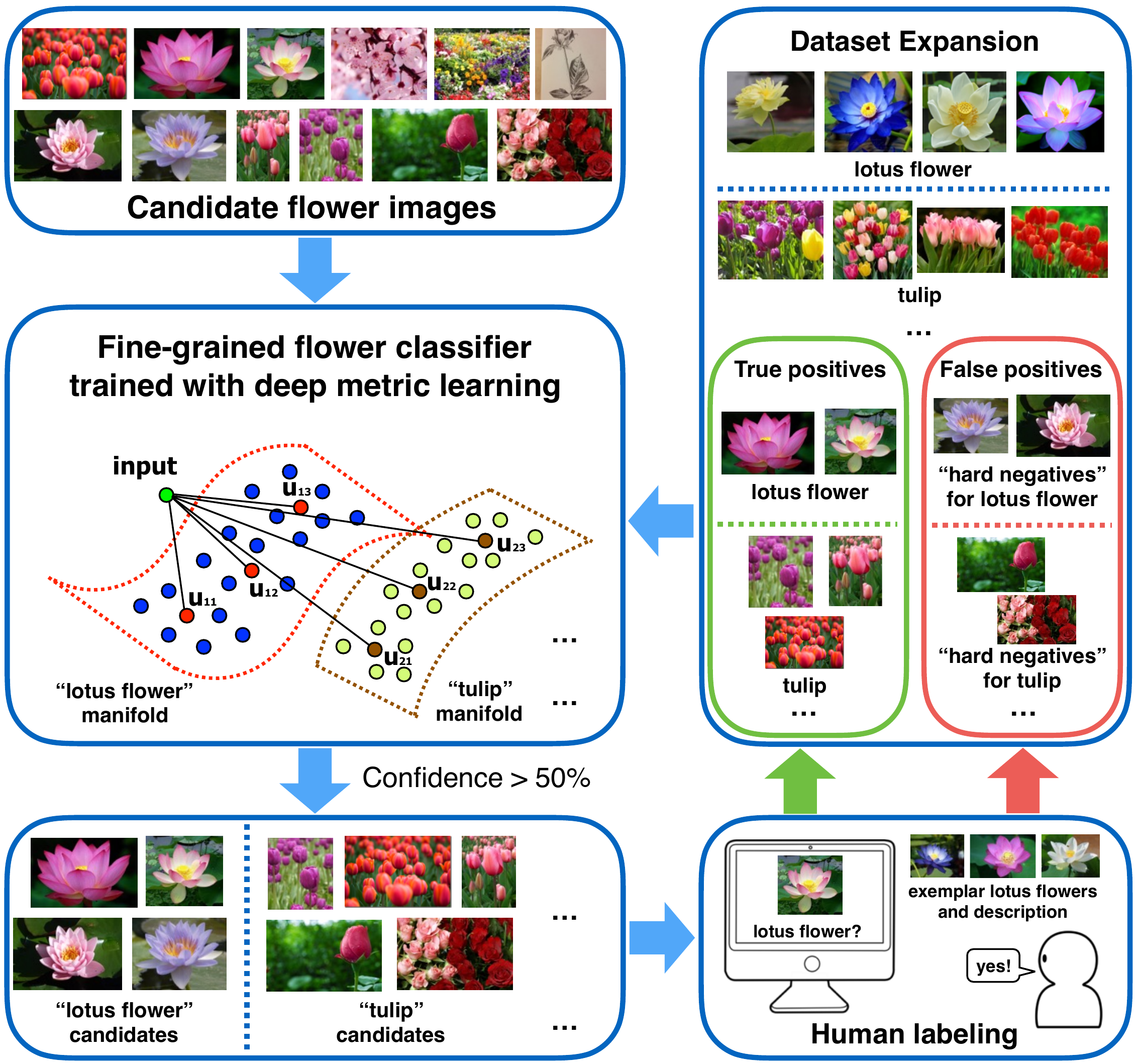}
\end{center}
\vspace{-2mm}
\caption{Overview of the proposed framework. Using deep metric learning with humans in the loop, we learn a low dimensional feature embedding for each category that can be used for fine-grained visual categorization and iterative dataset bootstrapping.}
\label{fig:overview}
\vspace{-2mm}
\end{figure}

Most existing FGVC methods fall into a classical two-step scheme: feature extraction followed by classification \cite{angelova2013efficient, berg2013poof, chai2013symbiotic, qian2015fine}.
Since these two steps are independent, the performance of the whole system is often suboptimal compared with an end-to-end system using Convolutional Neural Networks (CNN) that can be globally optimized via back-propagation \cite{pose_normlized_net, zhang2014part, krause2015fine, lin2015bilinear}.
Therefore, in this work, we focus on developing an end-to-end CNN-based method for FGVC.
However, compared with general purpose visual categorization, there are three main challenges arising when using such end-to-end CNN-based systems for FGVC.

Firstly, \textbf{lack of training data}. 
Current commonly used CNN architectures such as AlexNet \cite{alexnet}, VGGNet \cite{vggnet}, GoogLeNet-Inception \cite{googlenet} and ResNet \cite{resnet} have large numbers of parameters that require vast amounts of training data to achieve reasonably good performance.
Commonly used FGVC databases \cite{flower_102, branson2010visual, stanford_dog, stanford_car}, however, are relatively small, typically with less than a few tens of thousands of training images.

Secondly, compounding the above problem, FGVC can involve \textbf{large numbers of categories}.
For example, arguably, it is believed that there are more than $400,000$ species of flowers in the world \cite{joppa2011many}.
As a point of reference, modern face identification systems need to be trained on face images coming from millions of different identities (categories).
In such scenarios, the final fully connected layer of a CNN before the softmax layer would contain too many nodes, thereby making the training infeasible.

\begin{figure}[t]
\begin{center}
\includegraphics[width=0.9\linewidth]{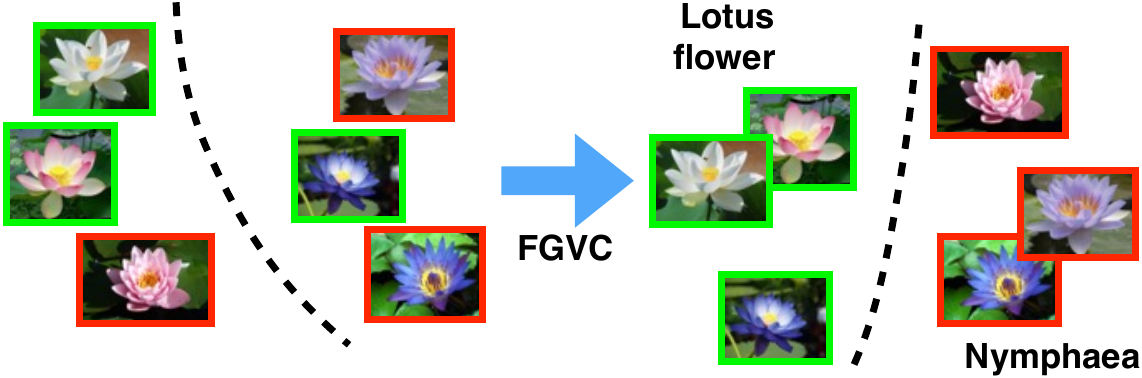}
\end{center}
\caption{Simple appearance based methods will likely find incorrect groups for two visually similar categories. A successful FGVC approach should be able to deal with the challenge of high intra-class vs.\ low inter-class variance.}
\label{fig:flower}
\vspace{-2mm}
\end{figure}

Lastly, \textbf{high intra-class vs.\ low inter-class variance}.
In FGVC, we confront two somewhat conflicting requirements: distinguishing visually similar images from different categories while allowing reasonably large variability (pose, color, lighting conditions, etc.) within a category.
As an example illustrated in Fig.\ \ref{fig:flower}, images from different categories could have similar shape and color.
On the other hand, sometimes images within same category can be very dissimilar due to nuisance variables. 
In such a scenario, since approaches that work well on generic image classification often focus on inter-class differences rather than intra-class variance, directly applying them to FGVC could make visually similar categories hard to be distinguished.

In this paper, we propose a framework that aims to address all three challenges.
We are interested in the following question: given an FGVC task with its associated training and test set, are we able to improve the performance by bootstrapping more training data from the web?
In light of this, we propose a unified framework using deep metric learning with humans in the loop, illustrated in Fig.\ \ref{fig:overview}.

We use an iterative approach for dataset bootstrapping and model training.
In each round, the model trained from last round is used to generate fine-grained confidence scores (probability distribution) for all the candidate images on categories.
Only images with highest confidence score larger than a threshold are kept and put into the corresponding category.
Then, for each category, by comparing with exemplar images and category definitions, human labelers remove false positives (hard negatives).
Images that pass the human filtering will be included into the dataset as new (vetted) data. 
Finally, we re-train our classification model by incorporating newly added data and also leveraging the hard negatives marked by human labelers.
The updated model will be used for the next round of dataset bootstrapping.
Although we focus on flower categorization in this work, the proposed framework is applicable to other FGVC tasks.

In order to capture within-class variance and utilize hard negatives as well, we propose a triplet-based deep metric learning approach for model training.
A novel metric learning approach enables us to learn low-dimensional manifolds with multiple anchor points for each fine-grained category.
These manifolds capture within-category variances and remain discriminative to other categories.
The data can be embedded into a feature space with dimension much lower than the number of categories.
During the classification, we generate the categorical confidence score by using multiple anchor points located on the manifolds.

In summary, the proposed framework handles all three challenges in FGVC mentioned above.
Using the proposed framework, we are able to grow our training set and get a better fine-grained classifier as well. 


\section{Related Work}
\label{sec:relatedwork}

\textbf{Fine-Grained Visual Categorization (FGVC)}.
Many approaches have been proposed recently for distinguishing between fine-grained categories.
Most of them \cite{angelova2013efficient, berg2013poof, chai2013symbiotic, qian2015fine} use two independent steps: feature extraction and classification.
Fueled by the recent advances in Convolutional Neural Networks (CNN) \cite{alexnet, r-cnn}, researchers have gravitated to CNN features \cite{pose_normlized_net, zhang2014part, krause2015fine, qian2015fine, lin2015bilinear} rather than traditional hand-crafted features such as LLC \cite{LLC_feature} or Fisher Vectors \cite{FV_feature}.
Sometimes, the information from segmentation \cite{krause2015fine}, part annotations \cite{pose_normlized_net}, or both \cite{chai2013symbiotic} is also used during the feature extraction.
Current state-of-the-art methods \cite{pose_normlized_net, zhang2014part, krause2015fine, lin2015bilinear} all adopt CNN-based end-to-end schemes that learn feature representations from data directly for classification. 
Although our method also draws upon a CNN-based scheme, there are two major differences. 1) Rather than using softmax loss, we aim to find a low-dimensional feature embedding for classification. 2) We incorporate humans into the training loop, with the human-provided input contributing to the training of our model.

\textbf{Fine-Grained Visual Datasets}.
Popular fine-grained visual datasets \cite{flower_102, ucsd_bird, stanford_dog, stanford_car} are relatively small scale, typically consisting of around 10 thousand training images or less.
There are some efforts recently in building large-scale fine-grained datasets \cite{merlin_bird, cuhk_car}.
We differ from these efforts both in terms of our goal and our approach.
Instead of building a dataset from scratch, we aim to bootstrap more training data to enlarge the existing dataset we have.
In addition, instead of human labeling, we also use a classifier to help during the dataset bootstrapping.
The most similar work in terms of dataset bootstrapping comes from Yu et al.\ \cite{lsun}, which builds a large-scale scene dataset with $10$ common categories using deep learning with humans in the loop.
However, we are bootstrapping a fine-grained dataset with much more categories (620). Moreover, instead of a dataset, we can also get a model trained with combined human-machine efforts.

\textbf{Deep Metric Learning}.
Another line of related work is metric learning with CNNs using pairwise \cite{siamese, contrastive_loss} or triplet constraints \cite{deepranking, facenet, deep_metric_triplet}.
The goal is to use a CNN with either pairwise (contrastive) or triplet loss to learn a feature embedding that captures the semantic similarity among images.
Compared with traditional metric learning methods that rely on hand-crafted features \cite{xing2002distance, nca, weinberger2009metric, chechik2010ranking}, deep metric learning directly learns from data and achieves much better performance.
Recently, it has been successfully applied to variety of problems including face recognition and verification \cite{deepface, facenet}, image retrieval \cite{deepranking}, semantic hashing \cite{deep_hashing}, product design \cite{bell15siggraph}, geo-localization \cite{cvpr2015geolocalization} and style matching \cite{veit2015learning}.
In contrast with previous methods, we propose a novel strategy that enables the learning of continuous manifolds. In addition, we also bring humans in the loop and leverage their inputs during metric learning.

\section{Dataset Bootstrapping}
\label{sec:dataset}

One of the main challenges in fine-grained visual recognition is the scarcity of training data. 
Labeling of fine-grained categories is tedious because it calls for experts with specialized domain knowledge. 
This section presents a bootstrapping framework on how to grow a small scale, fine-grained dataset in an efficient manner.  


\subsection{Discovering Candidate Images}
\label{sec:candidate_images}

In this first step, we wish to collect a large pool of candidate images for fine-grained subcategories under a coarse category, \eg, flowers.
The most intuitive way to crawl images could resort to image search engines like Google or Bing.
However, those returned images are often iconic, presenting a single, centered object with a simple background, which is not representative of natural conditions.

On the other hand, with the prevalence of powerful personal cameras and social networks, people capture their day-to-day photos and share them via platforms like Instagram or Flickr.
Those natural images uploaded by web users offer us a rich source of candidate images, often with tags that hint at the semantic content.
So if we search ``flower" on Instagram, a reasonable portion of returned images should be flower images. Naturally, we will need a filtering process to exclude the non-flower images.

We first downloaded two million images tagged with ``flower" via the Instagram API.
To remove the images that clearly contain no flowers, we pre-trained a flower classifier based on GoogLeNet-Inception \cite{googlenet} with $70$k images.
By feeding all the downloaded images to this classifier, we retained a set of nearly one million images, denoted as $\mathcal{C}$, with confidence score larger than $0.5$.


\subsection{Dataset Bootstrapping with Combined Human-Machine Efforts}
\label{sec:dataset_bootstrapping}

Given an initial fine-grained dataset $\mathcal{S}_0$ of $N$ categories and a candidate set $\mathcal{C}$, the goal of dataset bootstrapping is to select a subset $\mathcal{S}$ of the images from $\mathcal{C}$ that match with the original $N$ categories.
We divided the candidate set into a list of $k$ subsets: $\mathcal{C} = \mathcal{C}_1 \cup \mathcal{C}_2 \cup \dots \cup \mathcal{C}_k$ and used an iterative approach for dataset bootstrapping with $k$ iterations in total. 

Each iteration consists of three steps. 
Consider the $i$-th iteration.
First, we trained a CNN-based classifier (see Sec.\ \ref{sec:metric}) using the seed dataset $\mathcal{S}_{i-1} \cup \mathcal{H}_{i-1}$, where $\mathcal{H}_{i-1}$ contains the hard negatives from the previous step.
Second, using this classifier, we assigned each candidate image $x \in \mathcal{C}_i$ to one of the $N$ categories. 
Images with confidence score larger than $0.5$ form a high quality candidate set $\mathcal{D}_i \subset \mathcal{C}_i$ for the original $N$ categories. 
Third, we asked human labelers with domain expertise to identify true positives $\mathcal{T}_i$ and false positives $\mathcal{F}_i$, where $\mathcal{T}_i \cup \mathcal{F}_i = \mathcal{D}_i$.
Exemplar images and category definitions were shown to the labelers.

Compared to the traditional process requiring the labeler to select one of $N$ categories per image, we asked labelers to focus on a binary decision task which entails significantly less cognitive load. 
Noting that these false positives $\mathcal{F}_i$ are very similar to ground-truths, we regard them as hard negatives $\mathcal{H}_i \leftarrow \mathcal{H}_{i-1} \cup \mathcal{F}_i$.
True positives were also included to expand our dataset: $\mathcal{S}_{i} \leftarrow \mathcal{S}_{i-1} \cup \mathcal{T}_i$ for the next iteration.

It is worth mentioning this bootstrapping framework is similar in spirit to the recent work \cite{VondrickPR13,HeilbronEGN15} that used semi-automatic crowdsourcing strategy to collect and annotate videos. However, the key difference is we design a deep metric learning method (see Sec.\ \ref{sec:metric}) that specifically makes the use of the large number of hard negatives $\mathcal{H}_i$ in each iteration.




\section{Deep Metric Learning for FGVC}
\label{sec:metric}

We frame our problem as a deep metric learning task.
We choose metric learning for mainly two reasons.
First, compared with classic deep networks that use softmax loss in training, metric learning enables us to find a low-dimensional embedding that can well capture high intra-class variance.
Second, metric learning is a good way to leverage human-labeled hard negatives.
It is often difficult to get categorical labels for these hard negatives.
They could belong to flower species outside the dataset, or non-flower images.
Therefore, directly incorporating human-labeled hard negatives into a multi-way classification scheme such as softmax is infeasible, while it is quite natrual to include them into the metric learning.

\begin{figure}[t]
\begin{center}
\includegraphics[width=0.9\linewidth]{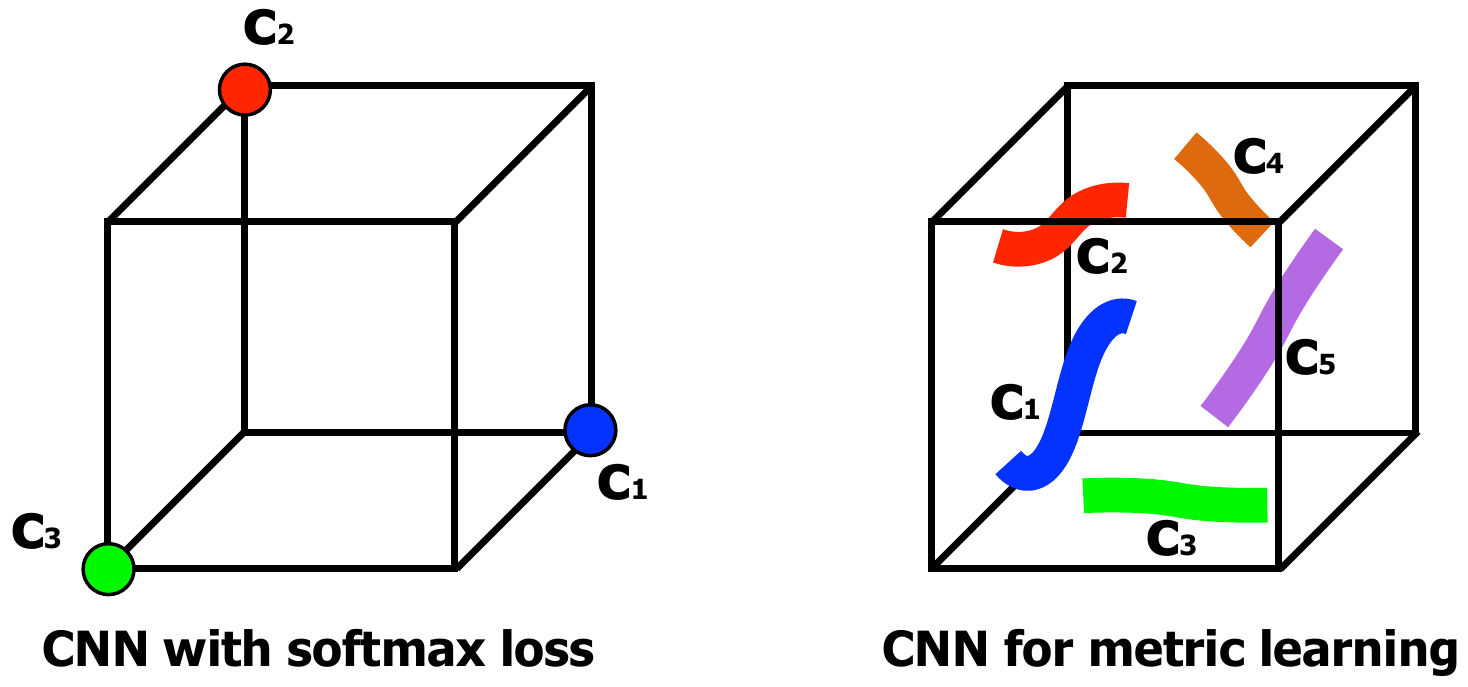}
\end{center}
\vspace{-2mm}
\caption{Comparison between CNN with softmax and CNN for metric learning in feature space, where $c_i$ denotes a group of images within the same category.}
\label{fig:comparison}
\vspace{-2mm}
\end{figure}

Fig.\ \ref{fig:comparison} illustrates the differences between CNN with softmax and CNN for metric learning in 3-dimensional feature space.
In order to minimize softmax loss, we try to map all images within the same category to a single point in feature space, which loses the intra-class variance.
In this figure, we try to map category $c_1$ to $[1, 0, 0]^\top$, $c_2$ to $[0, 1, 0]^\top$ and $c_3$ to $[0, 0, 1]^\top$, respectively.
We need $N$ nodes in final feature layer to represent $N$ categories.
However, in metric learning, we can learn manifolds and the dimensionality of the feature layer could be much smaller than $N$. In addition, the manifold can preserve useful intra-class variances such as color and pose.

Our goal is to learn a non-linear low-dimensional feature embedding $f(\cdot)$ via CNN, such that given two images $x$ and $y$, the Euclidean distance between $f(x)$ and $f(y)$ can reflect their semantic dissimilarity (whether they come from same category or not).
Typically, people use pairwise or triplet information to learn the feature embedding.

In the pairwise case \cite{siamese, contrastive_loss}, $f(\cdot)$ is learned from a set of image pairs $\{ (x_i, y_i) \}$ with corresponding labels $\{ l_i \}$ indicating whether $x_i$ and $y_i$ is similar.
In the triplet case \cite{deepranking, deep_metric_triplet}, $f(\cdot)$ is learned from a set of image triplets $\{ (x, x_p, x_n) \}$, which constrains the reference image $x$ to be more similar with the image $x_p$ of the same category compared with any image $x_n$ of different class.
We can see triplet constraints offer more fine-grained information: by making use of relative comparisons it is adaptive to differing granularity of similarity while the pairwise counterpart is not.
We therefore use triplet information to develop an end-to-end CNN-based approach for FGVC.

\subsection{Triplet-based Deep Metric Learning}
\label{sec:triplet}
\begin{figure}[t]
\begin{center}
\includegraphics[width=1\linewidth]{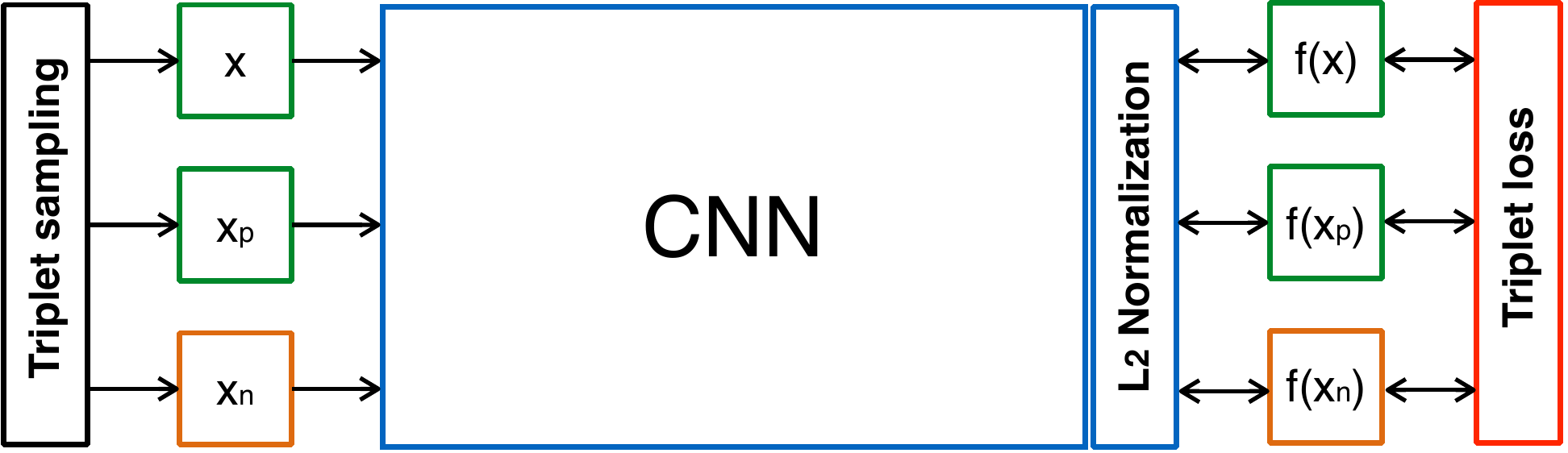}
\end{center}
   \caption{Triplet-based deep metric learning. In the input triplet, image $x$ is closer to $x_p$ than it is to $x_n$. We train a CNN to preserve this relative ordering under feature embedding $f(\cdot)$.}
\label{fig:triplet_framework}
\vspace{-2mm}
\end{figure}

The triplet-based deep metric learning framework is illustrated in Fig.\ \ref{fig:triplet_framework}.
In each iteration, the input triplet $(x, x_p, x_n)$ is sampled from the training set, where image $x$ is more similar to $x_p$ relative to $x_n$.
Then the triplet of three images are fed into an identical CNN simultaneously to get their non-linear feature embeddings $f(x)$, $f(x_p)$ and $f(x_n)$.
The CNN could be any arbitrary architecture such as AlexNet \cite{alexnet}, VGGNet \cite{vggnet} or GoogLeNet-Inception \cite{googlenet}.
Since we need to compute the distances in feature space, all the features should be normalized to eliminate the scale differences.
We use $L_2$-normalization for this purpose: $f(x) \leftarrow \frac{f(x)}{\sqrt{f(x)^\top f(x)}}$.

We use the triplet loss same as Wang et al.\ \cite{deepranking} used, which can be expressed as
\begin{equation}
\begin{aligned}
&\mathcal{L}_{triplet} (x, x_p, x_n) =\\
&\max\left\{ 0,  \left\lVert f(x) - f(x_p) \right\lVert_2^2 - \left\lVert f(x) - f(x_n) \right\lVert_2^2 + m\right\}
\end{aligned}
\label{eqn:triplet_loss}
\end{equation}
where $m$ is a hyper-parameter that controls the distance margin after the embedding.
This hinge loss function will produce a non-zero penalty of $\left\lVert f(x) - f(x_p) \right\lVert_2^2 - \left\lVert f(x) - f(x_n) \right\lVert_2^2 + m$ if the $L_2$ distance between $x$ and $x_n$ is smaller than the $L_2$ distance between $x$ and $x_p$ adding a margin $m$ in feature space: $\left\lVert f(x) - f(x_n) \right\lVert_2^2 < \left\lVert f(x) - f(x_p) \right\lVert_2^2 + m$.
The loss will be back propagated to each layer of the CNN and their corresponding parameters are updated through stochastic gradient descent.

\begin{figure*}[t]
\begin{center}
\includegraphics[width=0.95\linewidth]{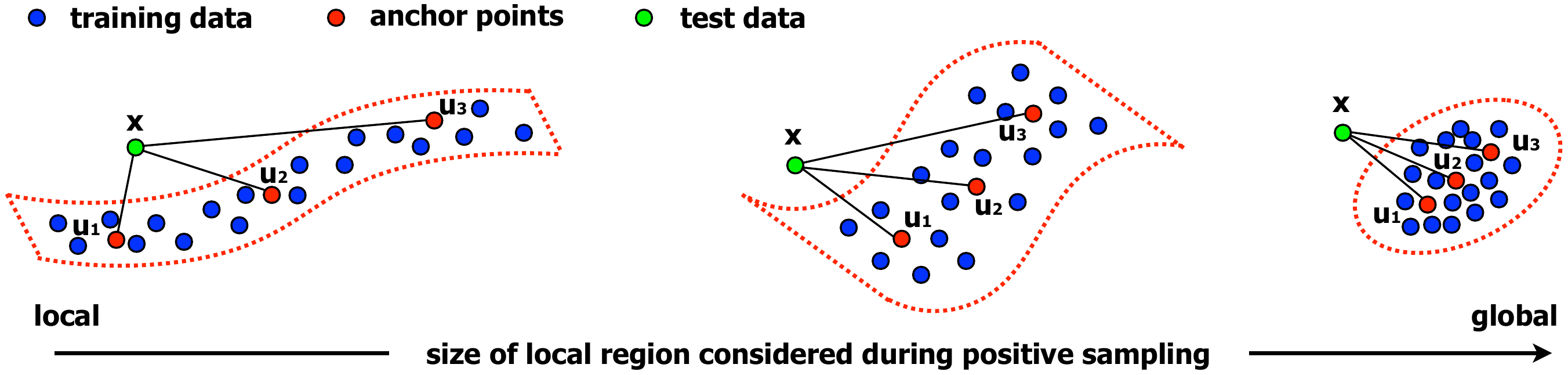}
\end{center}
   \caption{Control the shape of manifolds by sampling from local positive region. As the local region considered in positive sampling grows, the learned manifold will be increasingly dense.}
\label{fig:manifold}
\vspace{-3mm}
\end{figure*}

\subsection{Training from Hard Negatives}
\label{sec:hard_negative}
The most challenging part of training a triplet-based CNN lies in the triplet sampling.
Since there are $\mathcal{O}(n^3)$ possible triplets on a dataset with $n$ training data, going through all of them would be impractical for large $n$. 
A good triplet sampling strategy is needed to make training feasible.

We observed that during training, if we use randomly sampled triplets, many of them satisfy the triplet constraint well and give nearly zero loss in Eqn.\ \ref{eqn:triplet_loss}.
That is, those easy triplets have no effect in updating model parameters but we waste our time and resources in passing them through the network.
This makes the training process extremely inefficient and unstable: only few examples make contributions to the training within a batch.

Therefore, we use an online hard negatives mining scheme: only train on those triplets that violate the triplet constraint and give non-zero loss will be included into the training.
Why not simply train from the hardest negatives, i.e., triplets with the largest $\left\lVert f(x) - f(x_p) \right\lVert_2^2 - \left\lVert f(x) - f(x_n) \right\lVert_2^2$?
Because there are noisy data in the training set and trying to satisfy them ruins the overall performance.
A similar scenario was also reported in \cite{facenet}.

In our framework, instead of using images coming from categories that are different from the reference image, we also incorporate false positives marked by human labelers as hard negative candidates.
Those false positives are all misclassified by our model and thus provide us access to an excellent source of hard negatives.

\subsection{Learning Manifolds}
\label{sec:manifold}

Typically, given the reference image $x$, the positive image $x_p$ is sampled from all images within the same category as $x$.
Suppose we have a training set with $n$ images $\{x_i\}$ with labels $\{C(x_i)\}$ from $K$ categories, where $i = 1, 2, \dots, n$ and $C(x_i) \in \{1, 2, \dots, K\}$.
In this setting, considering a reference image $x$ within a fine-grained category, suppose the maximum between-class distance for $x$ in feature space is bounded by $D$. That is, $\left\lVert f(x) - f(x_n) \right\lVert_2 \leq D$, $\forall~ C(x_n) \neq C(x)$.
In order to have $0$ triplet loss for the reference image $x$, we need $\left\lVert f(x) - f(x_p) \right\lVert_2^2 \leq \left\lVert f(x) - f(x_n) \right\lVert_2^2 - m$, $\forall~ C(x_p) = C(x), C(x_n) \neq C(x)$.
Therefore, $\forall~ x_i, x_j$ where $C(x_i) = C(x_j) = C(x)$,
\begin{equation}
\begin{aligned}
\left\lVert f(x_i) - f(x_j) \right\lVert_2^2 &\leq \left\lVert f(x) - f(x_i) \right\lVert_2^2 + \left\lVert f(x) - f(x_j) \right\lVert_2^2\\
&\leq 2(D^2 - m)
\end{aligned}
\label{eqn:bound}
\end{equation}
The squared within-class pairwise distance is bounded by $2(D^2 - m)$.
Thus, by using triplet loss with positives sampled from all images in the same class, we are trying to map all images within that class into a hypersphere with radius $r = \frac{\sqrt{2(D^2 - m)}}{2}$.
In FGVC, between-class distances could be very small compared with the within-class distances.
In such a scenario, $D^2 - m$ could be very close to or even less than $0$, which makes the training process very difficult.

However, if we only force positives to be close to the reference locally, we are able to learn an extended manifold rather than a contracted sphere.
As illustrated in Fig.\ \ref{fig:manifold}, as the considered local positive region grows, the learned manifold will be increasingly contracted, eventually becoming a sphere when using all positives within the same category.

\begin{figure}[t]
\begin{center}
\includegraphics[width=0.9\linewidth]{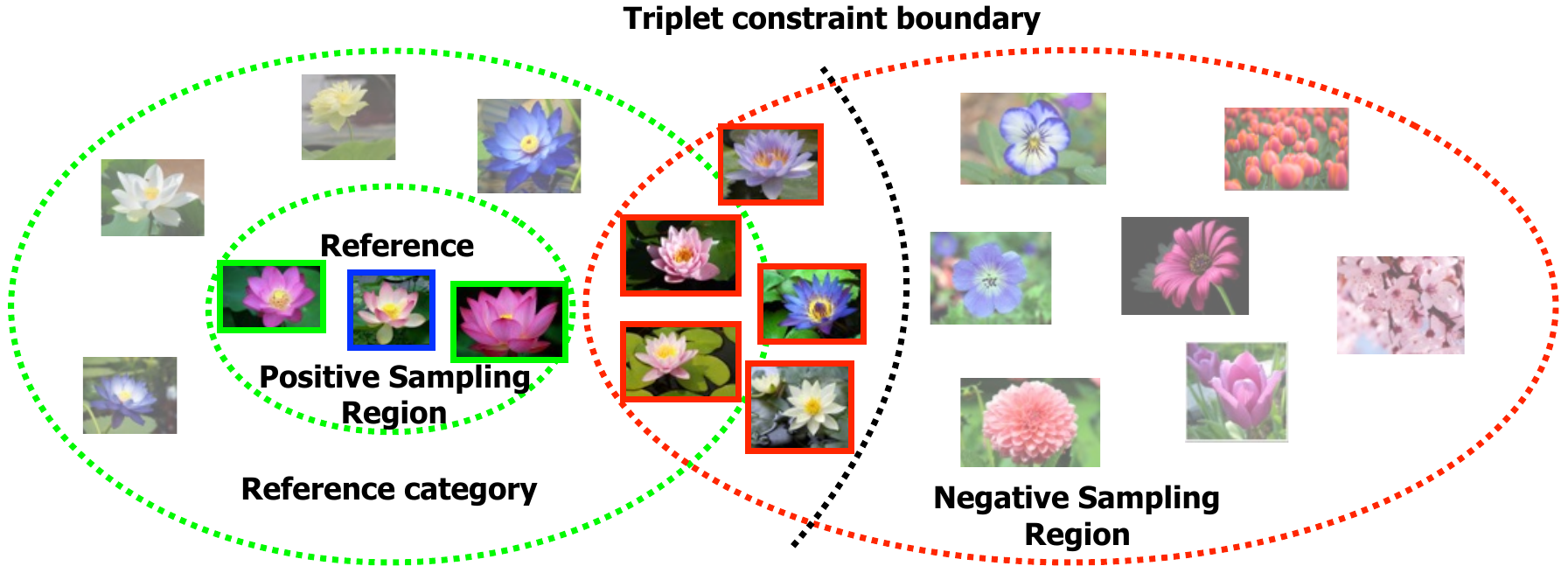}
\end{center}
   \caption{Triplet sampling strategy, in which for a reference image, positives are sampled locally and only hard negatives are kept.}
\label{fig:sampling}
\vspace{-2mm}
\end{figure}

The triplet sampling strategy we used is summarized in Fig.\ \ref{fig:sampling}.
Given a reference image $x$ (in the blue bounding box) we sample positive images $\{x_p\}$ (in the green bounding boxes) from the local region inside the same category.
Negative images $\{x_n\}$ are sampled from different categories but we only keep those hard negatives (marked by red bounding boxes): negatives that violate the triplet constraint with respect to the positives we chose.

\subsection{Classification}
\label{sec:classification}

After the manifold learning step, we adopt a soft voting scheme using anchor points on manifolds for classification.
For each category, the anchor points are generated by K-means clustering on the training set in feature space.
Suppose we have $N$ categories and each category has $K$ anchor points.
The $j$-th anchor point for category $i$ is represented as $u_{ij}$, where $i = 1, 2, \dots, N$, $j = 1, 2, \dots, K$.
Given an input query image $x$, we first extract its feature embedding $f(x)$ from our network, then the confidence score for category $i$ is generated as

\begin{equation}
p_i = \frac{\sum_{j=1}^K e^{ -\gamma  \left\lVert f(x) - u_{ij} \right\lVert_2^2 } }{\sum_{l=1}^N \big( \sum_{j=1}^k e^{ -\gamma  \left\lVert f(x) - u_{lj} \right\lVert_2^2 } \big)}
\label{eqn:soft_voting}
\end{equation}
The predicted label of $x$ is the category with the highest confidence score: $\arg\!\max_i p_i$.
$\gamma$ is a parameter controlling the ``softness" of label assignment and closer anchor points play more significant roles in soft voting.
If $\gamma \to \infty$, only the nearest anchor point is considered and the predicted label is ``hard" assigned to be the same as the nearest anchor point.
On the other hand, if $\gamma \to 0$, all the anchor points are considered to have the same contribution regardless of their distances between $f(x)$. 

Notice that during the prediction, the model is pre-trained offline and all the anchor points are calculated offline.
Therefore, given a query image, we only need a single forward pass in our model to extract the features.
Since we have learned a low-dimensional embedding, computing the distances between features and anchor points in low-dimensional space is very fast.

\subsection{Learning Anchor Points}
\label{sec:anchor_points}

\begin{figure}[t]
\begin{center}
\includegraphics[width=0.8\linewidth]{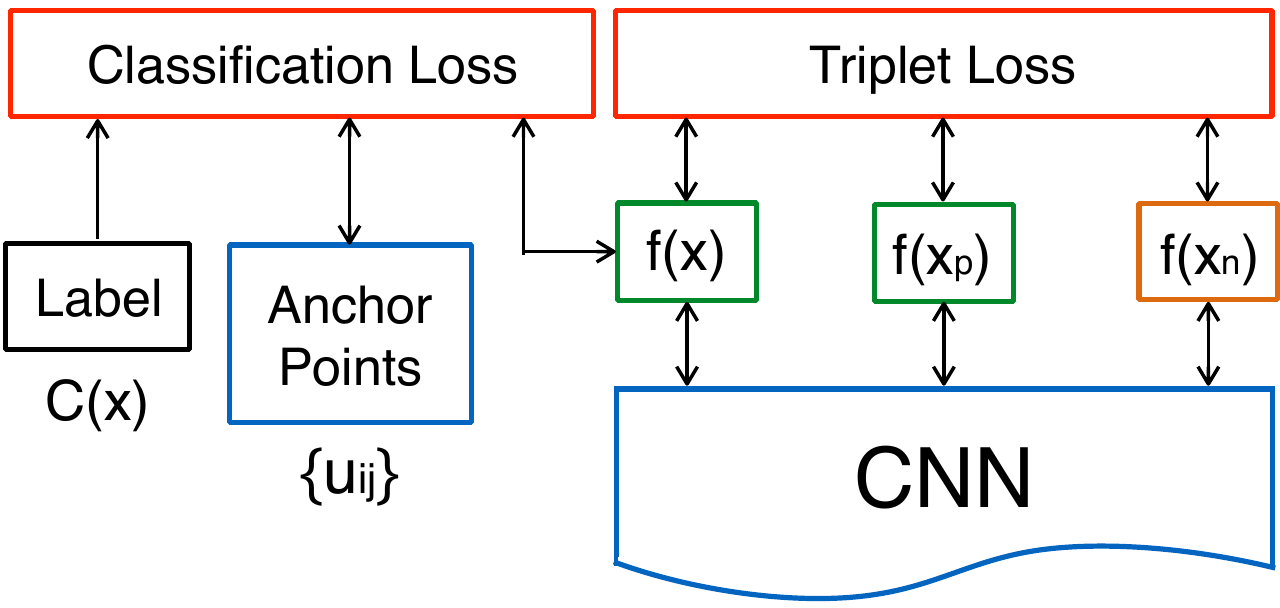}
\end{center}
   \caption{Combining anchor points learning into triplet network. The classification loss is used to update the anchor points.}
\label{fig:net}
\vspace{-2mm}
\end{figure}

As we just described, after metric learning, we use K-means to generate anchor points for representing manifolds and prediction. 
This could lead to suboptimal performance.
In fact, we can go one step further to directly learn anchor points by including soft voting into our triplet-based metric learning model, which is illustrated in Fig. \ref{fig:net}.
For simplicity, the data part is not shown.

In contrast to the previous model in Fig.\ \ref{fig:triplet_framework} that uses only triplet information, we also leverage the category label $C(x)$ for the reference image $x$ and learn anchor points for classification.
We can generate confidence scores $p_i$ for $f(x)$ using anchor points $\{u_{ij}\}$ by soft voting in Eqn.\ \ref{eqn:soft_voting}.
The classification loss we used is logistic loss on top of confidence score:
\begin{equation}
\mathcal{L}_{classification} (x, \{u_{ij}\}, C(x)) = - \log(p_{C(x)})
\label{eqn:classification_loss}
\end{equation}
where $p_{C(x)}$ is given in Eqn.\ \ref{eqn:soft_voting} by substituting $i$ with $C(x)$.
If we have very high confidence score on the true category, $p_{C(x)} \to 1$, then the loss will be very small: $\mathcal{L}_{classification} \to 0$.

The overall loss is the weighted sum of triplet and classification loss:

\begin{equation}
\mathcal{L} = \omega \mathcal{L}_{triplet} + (1 - \omega) \mathcal{L}_{classification}
\label{eqn:overall_loss}
\end{equation}

During training, the loss will be back-propagated to both CNN and anchor points.
Anchor point $u_{ij}$ will be updated based on the gradient of the loss with respect to $u_{ij}$: $\frac{\partial \mathcal{L}}{\partial u_{ij}}$.
Since we combine both triplet and categorical information and also learn anchor points directly for classification, we can expect better performance over the triplet-based model.

\section{Experimental Evaluation}
\label{sec:experiment}

In this section, we present experiments to evaluate the proposed deep metric learning approach against traditional two-step metric learning using deep features and commonly used softmax loss on our flower dataset and another publicly available dataset.
We also evaluate the effectiveness of dataset bootstrapping and training with humans in the loop.

\subsection{Experiments Setup}

We compare the performance of the proposed deep metric learning approach with the following baselines:
(1) Softmax loss for classification (\textbf{Softmax}). 
The most commonly used scheme in general purpose image classification. 
The deep network is trained from data with categorical label using softmax loss. 
We can get label prediction directly from the network output.
(2) Triplet loss with naive sampling (\textbf{Triplet-Naive}). 
The architecture illustrated in Fig.\ \ref{fig:triplet_framework} with randomly sampled triplets: given a reference image, the triplet is formed by randomly sampling a positive from same category and a negative from different category. 
Those triplets are directly fed into triplet network. During testing, we use the classification scheme described in Sec.\ \ref{sec:classification}.
(3) Triplet loss with hard negative mining (\textbf{Triplet-HN}). 
As discussed in Sec.\ \ref{sec:hard_negative}, instead of feeding all the triplets into the network, we only keep those hard negatives that violate triplet constraint.
(4) Triplet loss with manifold learning (\textbf{Triplet-M}). 
As mentioned in Sec.\ \ref{sec:manifold}, the positives are sampled locally with respect to the reference image from same category.
(5) Triplet loss with anchor points learning (\textbf{Triplet-A}). 
We combine anchor points learning with triplet network as illustrated in Fig.\ \ref{fig:net}. 
During testing, the network directly output label prediction based on confidence scores.
In addition, we also compared with state-of-the art FGVC approaches on publicly available dataset.

Since the network is trained via stochastic gradient descent, in order to do online sampling of triplets, we need to extract features on the entire training set, which is certainly inefficient if we do it for each iteration.
Therefore, as a trade-off, we adopt a quasi-online sampling strategy: 
after every $1,000$ iterations, we pause the training process and extract features on the training set, then based on their euclidean distances in feature space, we do triplet sampling (local positives and hard negatives) to generate a list of triplets for next $1,000$ iterations and resume the training process using the newly sampled triplets.

The CNN architecture we used is GoogLeNet-Inception \cite{googlenet}, which achieved state-of-the-art performance in large-scale image classification on ImageNet \cite{imagenet}.
All the baseline models are trained with fine-tuning using pre-trained GoogleNet-Inception on ImageNet dataset.

We used Caffe \cite{caffe}, an open source deep learning framework, for the implementation and training of our networks.
The models are trained on NVIDIA Tesla K80 GPUs.
The training process typically took about $5$ days on a single GPU to finish $200,000$ iterations with $50$ triplets in a batch per each iteration.

\subsection{Deep Metric Learning}

We evaluate the baselines on our flower dataset and publicly available CUB-200 Birds dataset \cite{ucsd_bird}.
There are several parameters in our model and the best values are found through cross-validation.
For all the following experiments on both dataset, we set the margin $m$ in triplet loss to be $0.2$; the feature dimension for $f(\cdot)$ to be $64$; the number of anchor points per each category $K$ to be $3$; the $\gamma$ in soft voting to be $5$.
We set $\omega = 0.1$ to make sure that the triplet loss term and the classification loss term in Eqn.\ \ref{eqn:overall_loss} have comparable scale.
For the size of positive sampling region, we set it to be $60\%$ of nearest neighbors within same category.
The effect of positive sampling region size will also be presented later in this section.

\textbf{Flowers-620.}
\textit{flowers-620} is the dataset we collected and used for dataset bootstrapping, which contains $20,211$ images from $620$ flower species, in which $15,437$ images are used for training.
The performance comparison of mean accuracy is summarized in Tab.\ \ref{tab:flower}.

\begin{table}[h]
\centering
\begin{tabular}{ |c|c| } 
 \hline
 Method (feature dimension) & Accuracy (\%) \\ 
 \hline
 \hline
 Softmax (620) & 65.1 \\ 
 \hline
 \hline
 Triplet-Naive (64) & 48.7 \\ 
 \hline
 Triplet-HN (64) & 64.6 \\ 
 \hline
 Triplet-M (64) & 65.9 \\ 
 \hline
 Triplet-A (64) & \textbf{66.8} \\ 
 \hline
\end{tabular}
\vspace{2mm}
\caption{Performance comparison on our \textit{flowers-620} dataset.}
\label{tab:flower}
\vspace{-2mm}
\end{table}

From the results, we have the following observations:
(1) Triplet-Naive, which uses randomly offline sampling, performed much worse compared with other triplet baselines, which clearly shows the importance of triplet sampling in training.
(2) Accuracy increases from Triplet-HN to Triplet-M, showing the effectiveness of learning a better manifolds with local positive sampling.
(3) Triplet-A performed best and achieved higher accuracy than Softmax.
This verifies our intuition that fine-grained categories often have high intra-class difference and such within-class variance can be well captured by learning manifolds with multiple anchor points.
In this way, even in a much lower dimensional feature space, the discrimination of the data can still be well preserved.
While in Softmax, we are trying to map all the data within a category to a single point in feature space, which fails to capture the within-class structure well.

\textbf{Birds-200.}
\textit{birds-200} is the Caltech-UCSD Birds-200-2011 data set for fine-grained birds categorization.
There are $11,788$ images from $200$ bird species. Each category has around $30$ images for training. 
In training and testing, we use the ground truth bounding boxes to crop the images before feeding them to the network.
The performance comparison is summarized in Tab.\ \ref{tab:bird}.

\begin{table}[h]
\centering
\begin{tabular}{ |c|c| } 
 \hline
 Method (feature dimension) & Accuracy (\%) \\ 
 \hline
 \hline
 Alignments \cite{gavves2014local} & 67.0 \\
 \hline
 MsML \cite{qian2015fine} & 67.9 \\ 
 \hline
 Symbiotic* \cite{chai2013symbiotic} & 69.5 \\
 \hline
 POOF* \cite{berg2013poof} & 73.3 \\
 \hline
 PB R-CNN* \cite{zhang2014part} & 82.0 \\
 \hline
 B-CNN \cite{lin2015bilinear} & 85.1 \\
 \hline
 PNN* \cite{pose_normlized_net} & 85.4 \\
 \hline
 \hline
 Softmax (620) & 77.2 \\ 
 \hline
 \hline
 Triplet-Naive (64) & 61.2 \\ 
 \hline
 Triplet-HN (64) & 77.9 \\ 
 \hline
 Triplet-M (64) & 79.3 \\ 
 \hline
 Triplet-A (64) & \textbf{80.7} \\ 
 \hline
\end{tabular}
\vspace{2mm}
\caption{Performance comparison on \textit{birds-200} dataset. ``*" indicates methods that use ground truth part annotations.}
\label{tab:bird}
\vspace{-2mm}
\end{table}

Similar to what we just observed in \textit{flowers-620}, experiment results verify the effectiveness of proposed methods.
We also compared to recent state-of-the-art approaches for fine-grained categorization.
Notice that we outperformed MsML \cite{qian2015fine} by a significant margin, which is a state-of-the-art metric learning method for FGVC.
Although our method performed worse than the recent proposed B-CNN \cite{lin2015bilinear}, we were able to achieve either better or comparable results with those state-of-the-arts using ground truth part annotations during training and testing.

We also evaluate the effect of local positive sampling region size.
As we mentioned earlier in Sec.\ \ref{sec:manifold}, the size of local positive sampling region controls the shape of manifolds.
We want to learn manifolds that can capture within-class variance well but not too spread out to lose the between-class discriminations.

\begin{figure}[t]
\centering
\subfigure[\textit{flowers-620}]{
\includegraphics[width=0.45\linewidth]{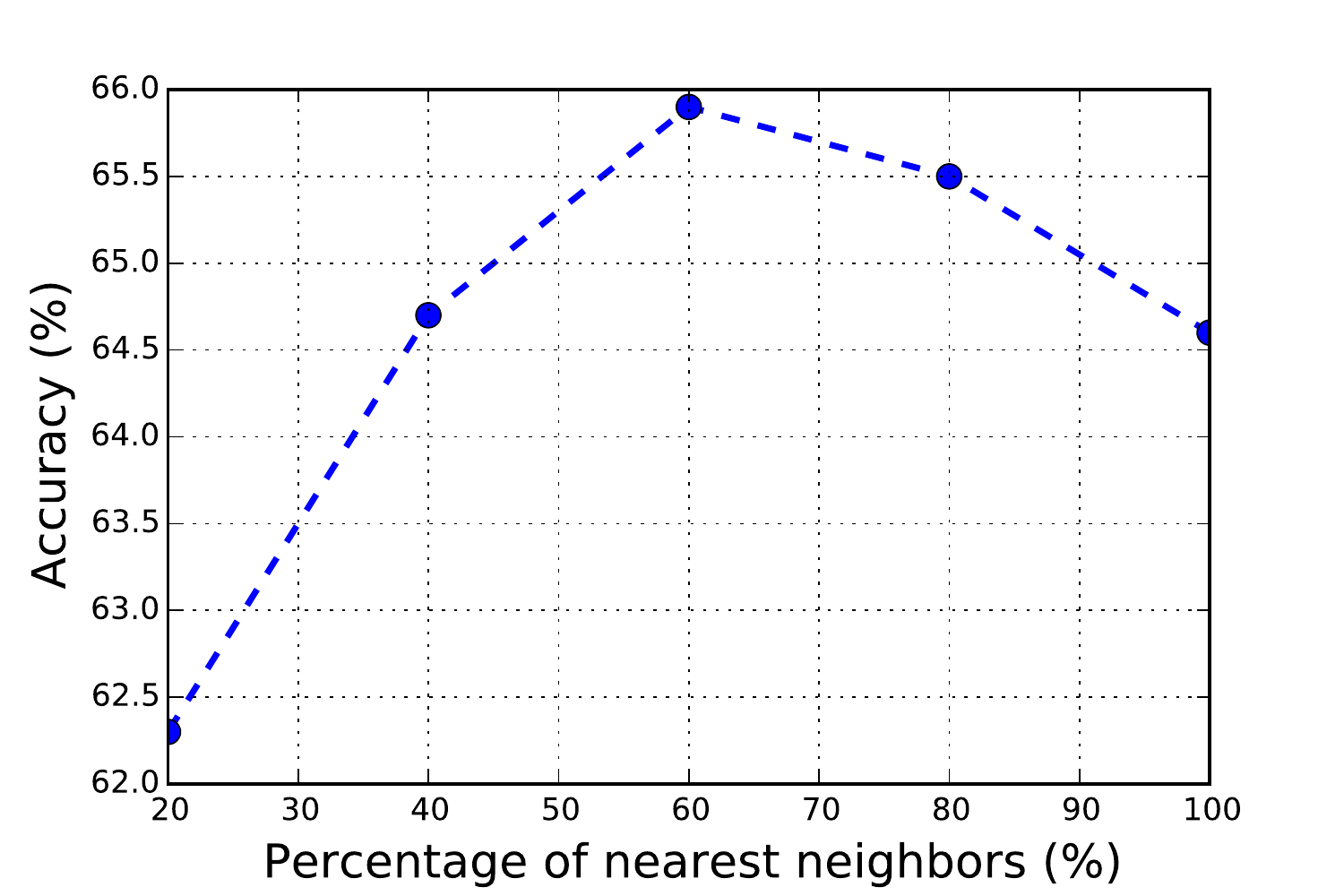}
   \label{fig:region_size_flower}
 }
\subfigure[\textit{birds-200}]{
\includegraphics[width=0.45\linewidth]{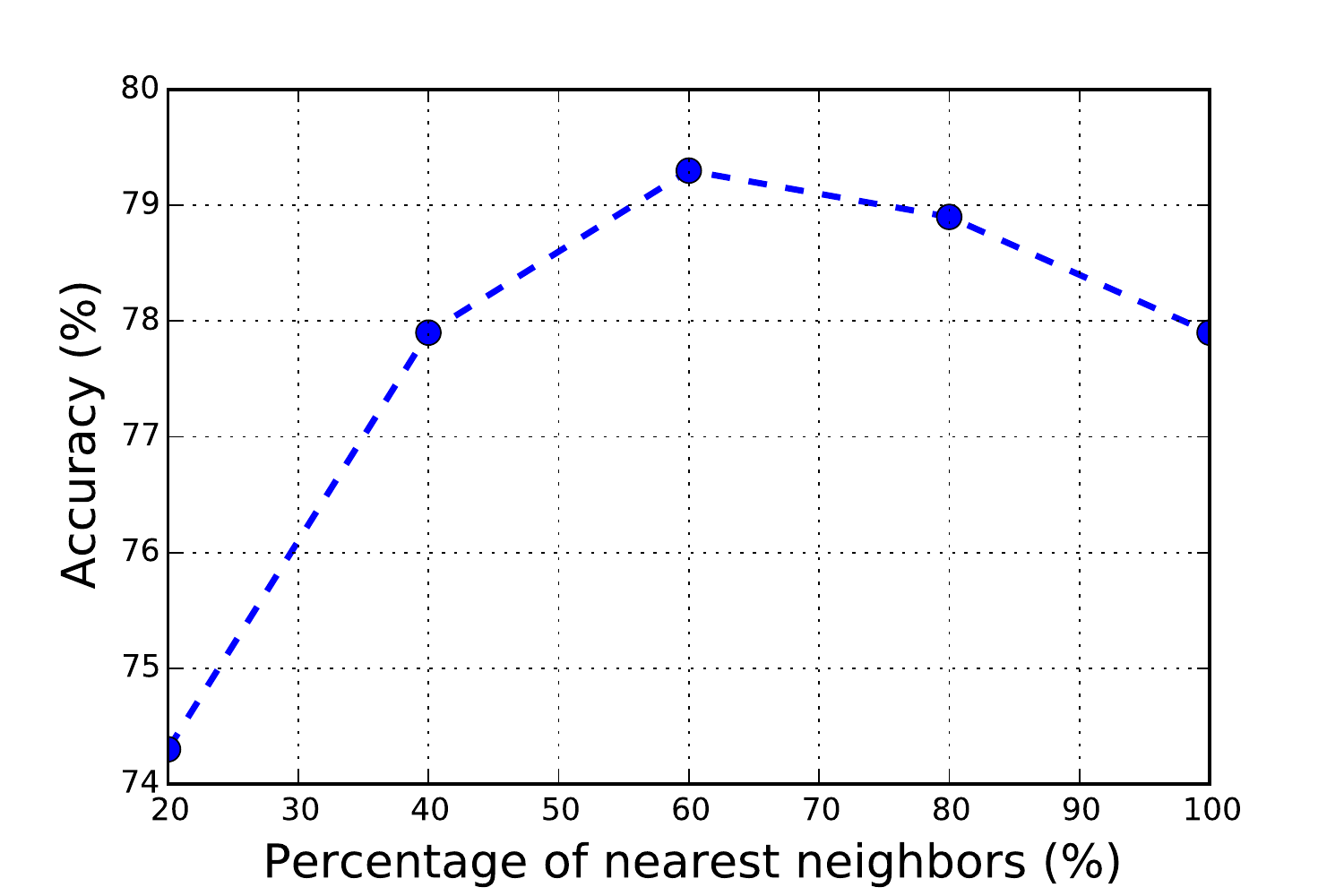}
   \label{fig:region_size_bird}
 }
\caption{Accuracy with varying positive sampling region size.}
\label{fig:region_size}
\end{figure}

Fig.\ \ref{fig:region_size} shows the mean accuracy with varying local positive sampling region using Triplet-M.
Using $60\%$ of nearest neighbors for positive sampling gives best results on both \textit{flowers-620} and \textit{birds-200}.

\subsection{Dataset Bootstrapping}

During dataset bootstrapping, other than true positives that passed human filtering and included into our dataset, plenty of false positives were marked by human labelers.
Those false positives are perfect hard negatives in our metric learning framework.
Therefore, we combined these human labeled hard negatives with negatives from other categories that violate triplet constraint during triplet sampling.
We sampled same number of human-labeled hard negatives as the hard negatives from other categories.

With the proposed framework, we included $11,567$ Instagram flower images into our database, which almost doubles the size of our training images to $27,004$.
At the same time, we also get $240,338$ hard negatives from labelers.
We call this new dataset \textit{flowers-620 + Ins} and will use it for the evaluation of dataset bootstrapping.
Notice that the test set in \textit{flowers-620 + Ins} remains same as \textit{flowers-620}.

For best quality, currently we only use in-house labelers.
Our framework could be deployed to crowdsourced labeling platforms like Amazon Mechanical Turk, bit with good quality control schemes.

We show that by dataset bootstrapping with humans in the loop, we are able to get a better model using the proposed metric learning approach.
For a fair comparison, we also include two baselines that enable hard negatives to be utilized in softmax scheme:
(1) SoftMax with all hard negatives as a single novel category (\textbf{Softmax + HNS}). 
The model is trained with one additional hard negative category. 
(2) SoftMax with hard negatives as multiple novel categories (\textbf{Softmax + HNM}). 
In this setting, instead of mixing all hard negatives as a single category, we regard hard negatives for different flower categories as different novel categories.
The model is trained with data from $620 \times 2 = 1240$ categories, from which $620$ of them are category-specific hard negatives.
To make the number of flower images and hard negatives to be balanced in each batch during training, the number of epochs we go through on all hard negatives is set to be $10\%$ of $620$ flower categories.
In testing, only confidence scores from $620$ flower categories will be considered for both baselines.
The experiment results on \textit{flowers-620 + Ins} are shown in Tab.\ \ref{tab:flower+ins}.

\begin{table}[t]
\centering
\begin{tabular}{ |c|c| } 
 \hline
 Method (feature dimension) & Accuracy (\%) \\ 
 \hline
 \hline
 Softmax (620) & 68.9 \\ 
 \hline
 Softmax + HNS (621) & 70.3 \\ 
 \hline
 Softmax + HNM (1240) & 70.8 \\ 
 \hline
 \hline
 Triplet-A (64) & 70.2 \\ 
 \hline
 Triplet-A + HN (64) & \textbf{73.7} \\ 
 \hline
\end{tabular}
\vspace{2mm}
\caption{Performance comparison on \textit{flowers-620 + Ins}.}
\vspace{-4mm}
\label{tab:flower+ins}
\end{table}

Compared with results in Tab.\ \ref{tab:flower}, we got $6.9\%$ improvement by dataset bootstrapping.
If we look at the breakdown, $3.4\%$ came from the newly added Instagram training images and $3.5\%$ came from human labeled hard negatives, indicating hard negatives has similar importance as positive images.
On the other hand, Softmax only gained $1.9\%$ by using hard negatives, which verifies our intuition that the triplet network is a better choice for utilizing hard negatives.
The proposed framework fully utilizes combined human-machine efforts to enlarge the dataset as well as train a better model.

\subsection{Visualization of Embedding}

\begin{figure}[t]
\begin{center}
\includegraphics[width=0.9\linewidth]{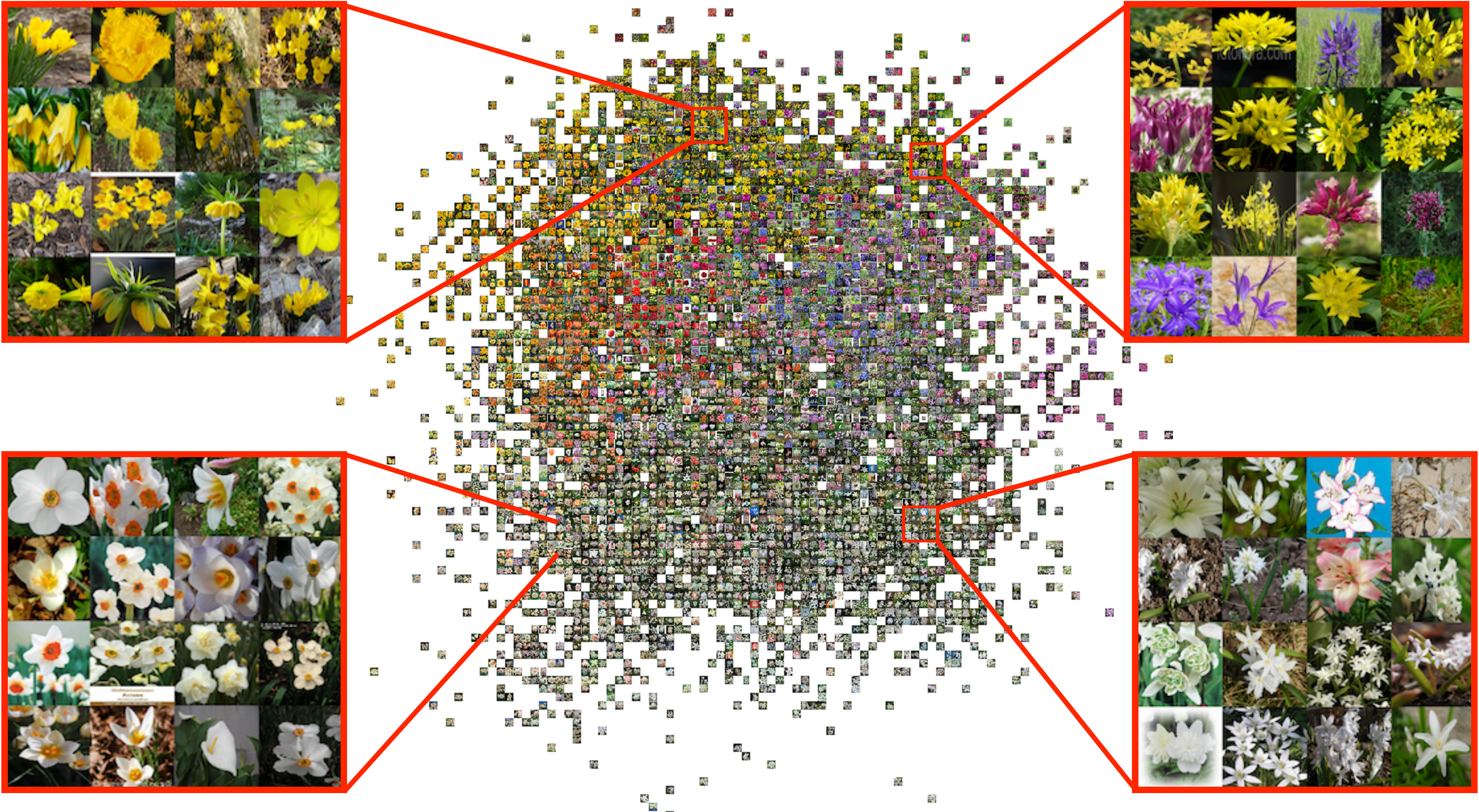}
\end{center}
   \caption{$2$-D embedding of \textit{flower-620} training set. We can observe that intra-class variance is captured in upper right and lower right regions.}
\label{fig:embedding}
\vspace{-2mm}
\end{figure}

For qualitative evaluation purpose, in Fig.\ \ref{fig:embedding}, we show the $2$-dimensional embedding of \textit{flower-620} training set using PCA on features extracted from the trained Triplet-A model.
Within the zoomed in regions, we can observe the effectiveness of our method in capturing high intra-class variances. 
For example, flowers from same category with different colors are mapped together in upper right and lower right regions.

\section{Conclusion and Discussion}
\label{sec:conlusion}
In this work, we have presented an iterative framework for fine-grained visual categorization and dataset bootstrapping based on a novel deep metric learning approach with humans in the loop.
Experimental results have validated the effectiveness of our framework.

We train our model mainly based on triplet information.
Although we adopt an effective and efficient online triplet sampling strategy, the training process could still be slow, which is a limitation of our method.
Some future work directions could be discovering and labeling novel categories during dataset bootstrapping with a combined human-machine framework or incorporating more information (\eg, hierarchical information, semantic similarity) into the triplet sampling strategy.

\newpage
{\small
\bibliographystyle{ieee}
\bibliography{ref.bib}

\begin{thebibliography}{10}\itemsep=-1pt

\bibitem{angelova2013efficient}
A.~Angelova and S.~Zhu.
\newblock Efficient object detection and segmentation for fine-grained
  recognition.
\newblock In {\em CVPR}, 2013.

\bibitem{LLC_feature}
A.~Angelova and S.~Zhu.
\newblock Efficient object detection and segmentation for fine-grained
  recognition.
\newblock In {\em CVPR}, 2013.

\bibitem{flower_nec}
A.~Angelova, S.~Zhu, and Y.~Lin.
\newblock Image segmentation for large-scale subcategory flower recognition.
\newblock In {\em WACV}, 2013.

\bibitem{bell15siggraph}
S.~Bell and K.~Bala.
\newblock Learning visual similarity for product design with convolutional
  neural networks.
\newblock {\em ACM Trans. on Graphics}, 2015.

\bibitem{berg2013poof}
T.~Berg and P.~N. Belhumeur.
\newblock Poof: Part-based one-vs.-one features for fine-grained
  categorization, face verification, and attribute estimation.
\newblock In {\em CVPR}, 2013.

\bibitem{pose_normlized_net}
S.~Branson, G.~Van~Horn, P.~Perona, and S.~Belongie.
\newblock Improved bird species recognition using pose normalized deep
  convolutional nets.
\newblock In {\em BMVC}, 2014.

\bibitem{branson2010visual}
S.~Branson, C.~Wah, F.~Schroff, B.~Babenko, P.~Welinder, P.~Perona, and
  S.~Belongie.
\newblock Visual recognition with humans in the loop.
\newblock In {\em ECCV}. 2010.

\bibitem{chai2013symbiotic}
Y.~Chai, V.~Lempitsky, and A.~Zisserman.
\newblock Symbiotic segmentation and part localization for fine-grained
  categorization.
\newblock In {\em ICCV}, 2013.

\bibitem{chechik2010ranking}
G.~Chechik, V.~Sharma, U.~Shalit, and S.~Bengio.
\newblock Large scale online learning of image similarity through ranking.
\newblock {\em JMLR}, 2010.

\bibitem{siamese}
S.~Chopra, R.~Hadsell, and Y.~LeCun.
\newblock Learning a similarity metric discriminatively, with application to
  face verification.
\newblock In {\em CVPR}, 2005.

\bibitem{cui2014spatial}
Y.~Cui, Y.~Xiang, K.~Rong, R.~Feris, and L.~Cao.
\newblock A spatial-color layout feature for representing galaxy images.
\newblock In {\em WACV}, 2014.

\bibitem{imagenet}
J.~Deng, W.~Dong, R.~Socher, L.-J. Li, K.~Li, and L.~Fei-Fei.
\newblock Imagenet: A large-scale hierarchical image database.
\newblock In {\em CVPR}, 2009.

\bibitem{dieleman2015rotation}
S.~Dieleman, K.~W. Willett, and J.~Dambre.
\newblock Rotation-invariant convolutional neural networks for galaxy
  morphology prediction.
\newblock {\em Monthly Notices of the Royal Astronomical Society}, 2015.

\bibitem{FV_feature}
E.~Gavves, B.~Fernando, C.~G. Snoek, A.~W. Smeulders, and T.~Tuytelaars.
\newblock Fine-grained categorization by alignments.
\newblock In {\em ICCV}, 2013.

\bibitem{gavves2014local}
E.~Gavves, B.~Fernando, C.~G. Snoek, A.~W. Smeulders, and T.~Tuytelaars.
\newblock Local alignments for fine-grained categorization.
\newblock {\em International Journal of Computer Vision}, 111(2):191--212,
  2014.

\bibitem{r-cnn}
R.~Girshick, J.~Donahue, T.~Darrell, and J.~Malik.
\newblock Rich feature hierarchies for accurate object detection and semantic
  segmentation.
\newblock In {\em CVPR}, 2014.

\bibitem{nca}
J.~Goldberger, G.~E. Hinton, S.~T. Roweis, and R.~Salakhutdinov.
\newblock Neighbourhood components analysis.
\newblock In {\em NIPS}, 2004.

\bibitem{contrastive_loss}
R.~Hadsell, S.~Chopra, and Y.~LeCun.
\newblock Dimensionality reduction by learning an invariant mapping.
\newblock In {\em CVPR}, 2006.

\bibitem{resnet}
K.~He, X.~Zhang, S.~Ren, and J.~Sun.
\newblock Deep residual learning for image recognition.
\newblock {\em arXiv preprint arXiv:1512.03385}, 2015.

\bibitem{HeilbronEGN15}
F.~C. Heilbron, V.~Escorcia, B.~Ghanem, and J.~C. Niebles.
\newblock Activitynet: {A} large-scale video benchmark for human activity
  understanding.
\newblock In {\em CVPR}, 2015.

\bibitem{deep_metric_triplet}
E.~Hoffer and N.~Ailon.
\newblock Deep metric learning using triplet network.
\newblock {\em arXiv preprint arXiv:1412.6622}, 2014.

\bibitem{caffe}
Y.~Jia, E.~Shelhamer, J.~Donahue, S.~Karayev, J.~Long, R.~Girshick,
  S.~Guadarrama, and T.~Darrell.
\newblock Caffe: Convolutional architecture for fast feature embedding.
\newblock {\em arXiv preprint arXiv:1408.5093}, 2014.

\bibitem{joppa2011many}
L.~N. Joppa, D.~L. Roberts, and S.~L. Pimm.
\newblock How many species of flowering plants are there?
\newblock {\em Proceedings of the Royal Society B: Biological Sciences}, 2011.

\bibitem{stanford_dog}
A.~Khosla, N.~Jayadevaprakash, B.~Yao, and F.-F. Li.
\newblock Novel dataset for fgvc: Stanford dogs.
\newblock In {\em San Diego: CVPR Workshop on FGVC}, 2011.

\bibitem{krause2015fine}
J.~Krause, H.~Jin, J.~Yang, and L.~Fei-Fei.
\newblock Fine-grained recognition without part annotations.
\newblock In {\em CVPR}, 2015.

\bibitem{stanford_car}
J.~Krause, M.~Stark, J.~Deng, and L.~Fei-Fei.
\newblock 3d object representations for fine-grained categorization.
\newblock In {\em ICCVW}, 2013.

\bibitem{alexnet}
A.~Krizhevsky, I.~Sutskever, and G.~E. Hinton.
\newblock Imagenet classification with deep convolutional neural networks.
\newblock In {\em NIPS}, 2012.

\bibitem{leafsnap}
N.~Kumar, P.~N. Belhumeur, A.~Biswas, D.~W. Jacobs, W.~J. Kress, I.~C. Lopez,
  and J.~V. Soares.
\newblock Leafsnap: A computer vision system for automatic plant species
  identification.
\newblock In {\em ECCV}. 2012.

\bibitem{deep_hashing}
H.~Lai, Y.~Pan, Y.~Liu, and S.~Yan.
\newblock Simultaneous feature learning and hash coding with deep neural
  networks.
\newblock {\em arXiv preprint arXiv:1504.03410}, 2015.

\bibitem{larios2010haar}
N.~Larios, B.~Soran, L.~G. Shapiro, G.~Mart{\'\i}nez-Mu{\~n}oz, J.~Lin, and
  T.~G. Dietterich.
\newblock Haar random forest features and svm spatial matching kernel for
  stonefly species identification.
\newblock In {\em ICPR}, 2010.

\bibitem{cvpr2015geolocalization}
T.-Y. Lin, Y.~Cui, S.~Belongie, and J.~Hays.
\newblock Learning deep representations for ground-to-aerial geolocalization.
\newblock In {\em CVPR}, 2015.

\bibitem{lin2015bilinear}
T.-Y. Lin, A.~RoyChowdhury, and S.~Maji.
\newblock Bilinear cnn models for fine-grained visual recognition.
\newblock {\em arXiv preprint arXiv:1504.07889}, 2015.

\bibitem{flower_ox_1}
M.-E. Nilsback and A.~Zisserman.
\newblock A visual vocabulary for flower classification.
\newblock In {\em CVPR}, 2006.

\bibitem{flower_102}
M.-E. Nilsback and A.~Zisserman.
\newblock Automated flower classification over a large number of classes.
\newblock In {\em ICVGIP}, 2008.

\bibitem{qian2015fine}
Q.~Qian, R.~Jin, S.~Zhu, and Y.~Lin.
\newblock Fine-grained visual categorization via multi-stage metric learning.
\newblock In {\em CVPR}, 2015.

\bibitem{facenet}
F.~Schroff, D.~Kalenichenko, and J.~Philbin.
\newblock Facenet: A unified embedding for face recognition and clustering.
\newblock {\em arXiv preprint arXiv:1503.03832}, 2015.

\bibitem{vggnet}
K.~Simonyan and A.~Zisserman.
\newblock Very deep convolutional networks for large-scale image recognition.
\newblock {\em arXiv preprint arXiv:1409.1556}, 2014.

\bibitem{googlenet}
C.~Szegedy, W.~Liu, Y.~Jia, P.~Sermanet, S.~Reed, D.~Anguelov, D.~Erhan,
  V.~Vanhoucke, and A.~Rabinovich.
\newblock Going deeper with convolutions.
\newblock {\em arXiv preprint arXiv:1409.4842}, 2014.

\bibitem{deepface}
Y.~Taigman, M.~Yang, M.~Ranzato, and L.~Wolf.
\newblock Deepface: Closing the gap to human-level performance in face
  verification.
\newblock In {\em CVPR}, 2014.

\bibitem{merlin_bird}
G.~Van~Horn, S.~Branson, R.~Farrell, S.~Haber, J.~Barry, P.~Ipeirotis,
  P.~Perona, and S.~Belongie.
\newblock Building a bird recognition app and large scale dataset with citizen
  scientists: The fine print in fine-grained dataset collection.
\newblock In {\em CVPR}, 2015.

\bibitem{veit2015learning}
A.~Veit, B.~Kovacs, S.~Bell, J.~McAuley, K.~Bala, and S.~Belongie.
\newblock Learning visual clothing style with heterogeneous dyadic
  co-occurrences.
\newblock {\em arXiv preprint arXiv:1509.07473}, 2015.

\bibitem{VondrickPR13}
C.~Vondrick, D.~J. Patterson, and D.~Ramanan.
\newblock Efficiently scaling up crowdsourced video annotation - {A} set of
  best practices for high quality, economical video labeling.
\newblock {\em IJCV}, 101(1):184--204, 2013.

\bibitem{ucsd_bird}
C.~Wah, S.~Branson, P.~Welinder, P.~Perona, and S.~Belongie.
\newblock {The Caltech-UCSD Birds-200-2011 Dataset}.
\newblock Technical report, 2011.

\bibitem{deepranking}
J.~Wang, T.~Leung, C.~Rosenberg, J.~Wang, J.~Philbin, B.~Chen, Y.~Wu, et~al.
\newblock Learning fine-grained image similarity with deep ranking.
\newblock {\em arXiv preprint arXiv:1404.4661}, 2014.

\bibitem{weinberger2009metric}
K.~Q. Weinberger and L.~K. Saul.
\newblock Distance metric learning for large margin nearest neighbor
  classification.
\newblock {\em JMLR}, 2009.

\bibitem{xiang2015car}
Y.~Xiang, W.~Choi, Y.~Lin, and S.~Savarese.
\newblock Data-driven 3d voxel patterns for object category recognition.
\newblock In {\em CVPR}, 2015.

\bibitem{xing2002distance}
E.~P. Xing, M.~I. Jordan, S.~Russell, and A.~Y. Ng.
\newblock Distance metric learning with application to clustering with
  side-information.
\newblock In {\em NIPS}, 2002.

\bibitem{cuhk_car}
L.~Yang, P.~Luo, C.~C. Loy, and X.~Tang.
\newblock A large-scale car dataset for fine-grained categorization and
  verification.
\newblock In {\em CVPR}, 2015.

\bibitem{lsun}
F.~Yu, Y.~Zhang, S.~Song, A.~Seff, and J.~Xiao.
\newblock Construction of a large-scale image dataset using deep learning with
  humans in the loop.
\newblock {\em arXiv preprint arXiv:1506.03365}, 2015.

\bibitem{zhang2014part}
N.~Zhang, J.~Donahue, R.~Girshick, and T.~Darrell.
\newblock Part-based r-cnns for fine-grained category detection.
\newblock In {\em ECCV}. 2014.

\end{thebibliography}
}

\end{document}